\documentclass[journal]{IEEEtran}

\usepackage{amsmath}
\usepackage{amssymb}
\usepackage{mathabx}
\usepackage{mathrsfs}
\usepackage{scalerel,amssymb}

\usepackage{pifont}
\usepackage{wasysym}
\usepackage{tablefootnote}
\usepackage{tabu}
\usepackage[normalem]{ulem}
\usepackage[colorlinks,linkcolor={{blue!70!black}},pagebackref=true,breaklinks=true,colorlinks,bookmarks=false]{hyperref}
\usepackage{comment}
\usepackage{epsfig}
\usepackage{graphicx}
\usepackage{makecell}
\usepackage[font=bf,labelfont=bf]{caption}
\usepackage{subcaption}
\usepackage{soul}
\usepackage[table, svgnames]{xcolor}
\definecolor{Gray}{gray}{0.85}

\newcommand\blfootnote[1]{%
  \begingroup
  \renewcommand\thefootnote{}\footnote{#1}%
  \addtocounter{footnote}{-1}%
  \endgroup
}

\usepackage{multibib}
\newcites{latex}{\LaTeX-Literature}

\usepackage[utf8]{inputenc} 
\usepackage[T1]{fontenc}    
\usepackage{url}            
\usepackage{booktabs}       
\usepackage{amsfonts}       
\usepackage{nicefrac}       
\usepackage{microtype}      
\usepackage{lipsum}
\usepackage{tabularx}
\usepackage{tabu}

\usepackage{multirow}
\usepackage[]{algorithm2e}
\SetKwInput{KwInput}{Input}               
\SetKwInput{KwOutput}{Output}

\def\msquare{\mathord{\scalerel*{\Box}{gX}}}

\usepackage[english]{babel}
\addto\extrasenglish{%
}
\usepackage{amsthm}

\newcommand{\etal}{\textit{et al}. }
\newcommand{\ie}{\textit{i}.\textit{e}., }
\newcommand{\eg}{\textit{e}.\textit{g}. }
\newcommand{\x}{\mathbf{x}}

\newcommand{\vertex}{\mathcal{V}}
\newcommand{\edge}{\mathcal{E}}

\DeclareMathOperator{\argsort}{ARGSORT}
\DeclareMathOperator{\bool}{BOOLEAN}
\DeclareMathOperator{\select}{SELECT}
\DeclareMathOperator{\AGG}{AGG}

\DeclareMathOperator{\CAT}{CAT}

\usepackage{mathtools}
\usepackage{enumitem}

\newtheorem*{prop-non}{Proposition}

\begin{document}

\title{Position-Sensing Graph Neural Networks: \\
Proactively Learning Nodes Relative Positions
}

\author{Zhenyue~Qin\textsuperscript{*},
	Yiqun~Zhang\textsuperscript{*},
        Saeed~Anwar,
        Dongwoo~Kim\textsuperscript{\dag}, 
        Yang~Liu,
        Pan~Ji, 
        Tom~Gedeon\textsuperscript{\dag}
\thanks{Y. ZHANG, Z. QIN were with Australian National University (ANU). Y. LIU was with ANU and DATA61, CSIRO. S. ANWAR is with both KFUPM and SDAIA-KFUPM JRCAI. P. JI was with OPPO US Research Center. D. KIM was with POSTECH. T. GEDEON is with Curtin University, Obuda University Hungary, and ANU. Email: kf.zy.qin@gmail.com, dongwoo.kim@postech.ac.kr, tom.gedeon@curtin.edu.au. * indicates equal contribution. $\dag$ 
represents the corresponding author.}%
}

\markboth{}%
{Shell \MakeLowercase{\textit{et al.}}: Bare Demo of IEEEtran.cls for IEEE Journals}

\maketitle

\begin{abstract}
Most existing graph neural networks (GNNs) learn node embeddings using the framework of message passing and aggregation. Such GNNs are incapable of learning relative positions between graph nodes within a graph. To empower GNNs with the awareness of node positions, some nodes are set as anchors. Then, using the distances from a node to the anchors, GNNs can infer relative positions between nodes. 
However, P-GNNs arbitrarily select anchors, leading to compromising position-awareness and feature extraction. To eliminate this compromise, we demonstrate that selecting evenly distributed and asymmetric anchors is essential. On the other hand, we show that choosing anchors that can aggregate embeddings of all the nodes within a graph is NP-complete. Therefore, devising efficient optimal algorithms in a deterministic approach is practically not feasible. To ensure position-awareness and bypass NP-completeness, we propose Position-Sensing Graph Neural Networks (PSGNNs), learning how to choose anchors in a back-propagatable fashion. Experiments verify the effectiveness of PSGNNs against state-of-the-art GNNs, substantially improving performance on various synthetic and real-world graph datasets while enjoying stable scalability. 
Specifically, PSGNNs on average boost AUC more than 14\% for pairwise node classification and 18\% for link prediction over the existing state-of-the-art position-aware methods. 
Our source code is publicly available at: 
\href{https://github.com/ZhenyueQin/PSGNN}{https://github.com/ZhenyueQin/PSGNN}.

\end{abstract}

\begin{IEEEkeywords}
graph neural networks, position-aware
\end{IEEEkeywords}

\IEEEpeerreviewmaketitle

\section{Introduction}
A recent new trend of current studies on neural networks is to design graph neural networks (GNNs) for learning network structured data, such as molecules, biological and social networks~\cite{wu2020comprehensive,bacciu2020gentle}. To extract meaningful node representations, GNNs commonly follow a recursive neighborhood aggregation pattern. Specifically, each node aggregates its own and neighbors' feature vectors, followed by a non-linear transformation. As the iteration continues, a node steadily aggregates information of increasingly distant neighbors. We refer to this pattern as the approach of message passing and aggregating. This learning paradigm has been demonstrated to be effective on many tasks, such as classification for nodes and graphs, link prediction, and graph generation.   

Despite the effectiveness of learning node representations with the approach of message passing and aggregating, it cannot extract relative positions between nodes within a graph~\cite{you2019position}. However, knowing relative positions is essential for a variety of tasks~\cite{mohamed2019comprehensive,zhang2018link}. For example, nodes are more likely to have connections and locate in the same community with other nodes that have smaller distances between them~\cite{daud2020applications}. Thus, discovering relative distances between nodes can benefit tasks of link prediction and community detection. 

\begin{figure}[t]
\centering
\begin{subfigure}{.22\textwidth}
  \centering
  \includegraphics[width=.99\linewidth]{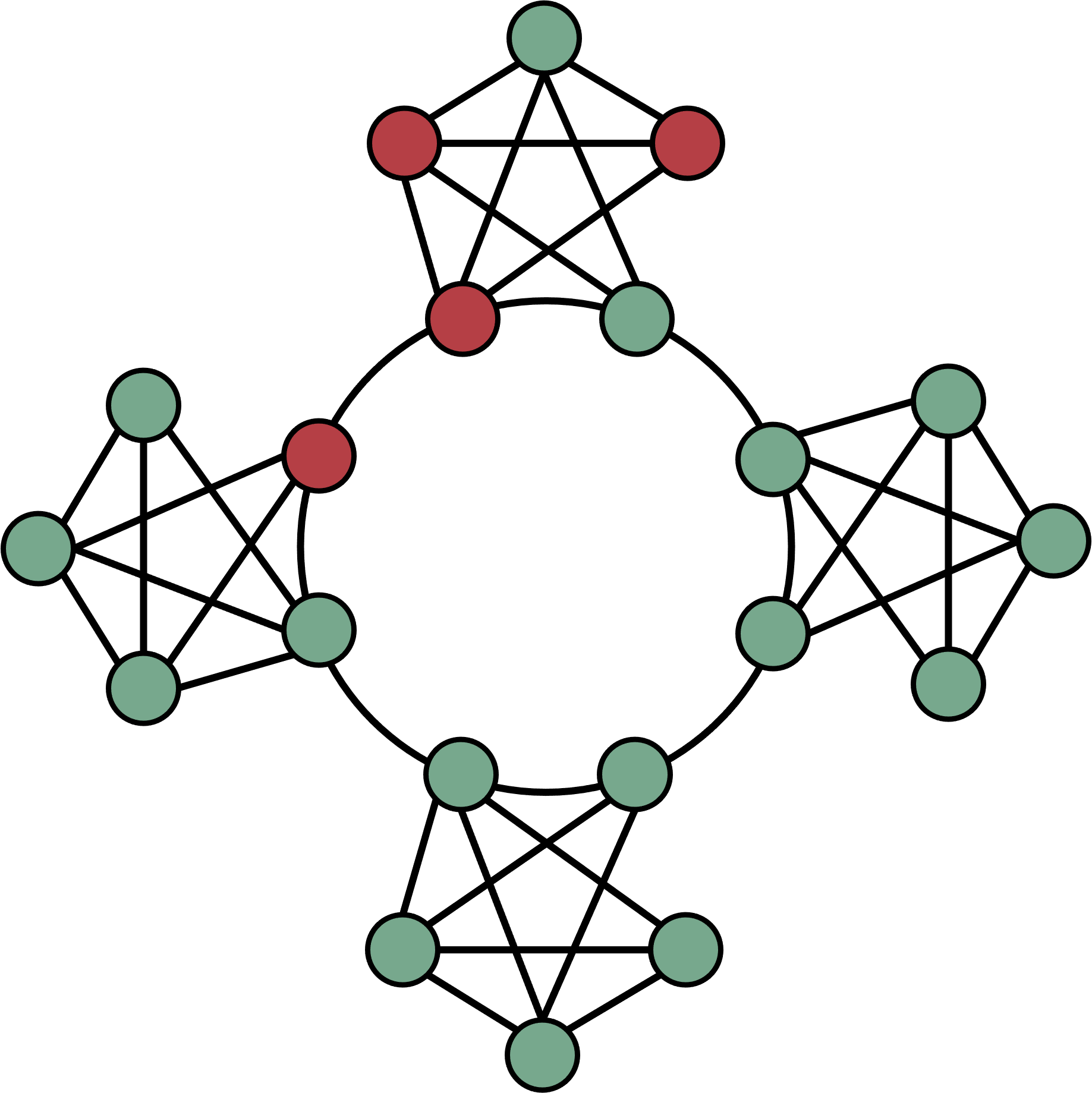} 
  \caption*{Anchors of P-GNN~\cite{you2019position}. }
\end{subfigure}
\hspace{0.3cm}
\begin{subfigure}{.22\textwidth}
  \centering
  \includegraphics[width=.99\linewidth]{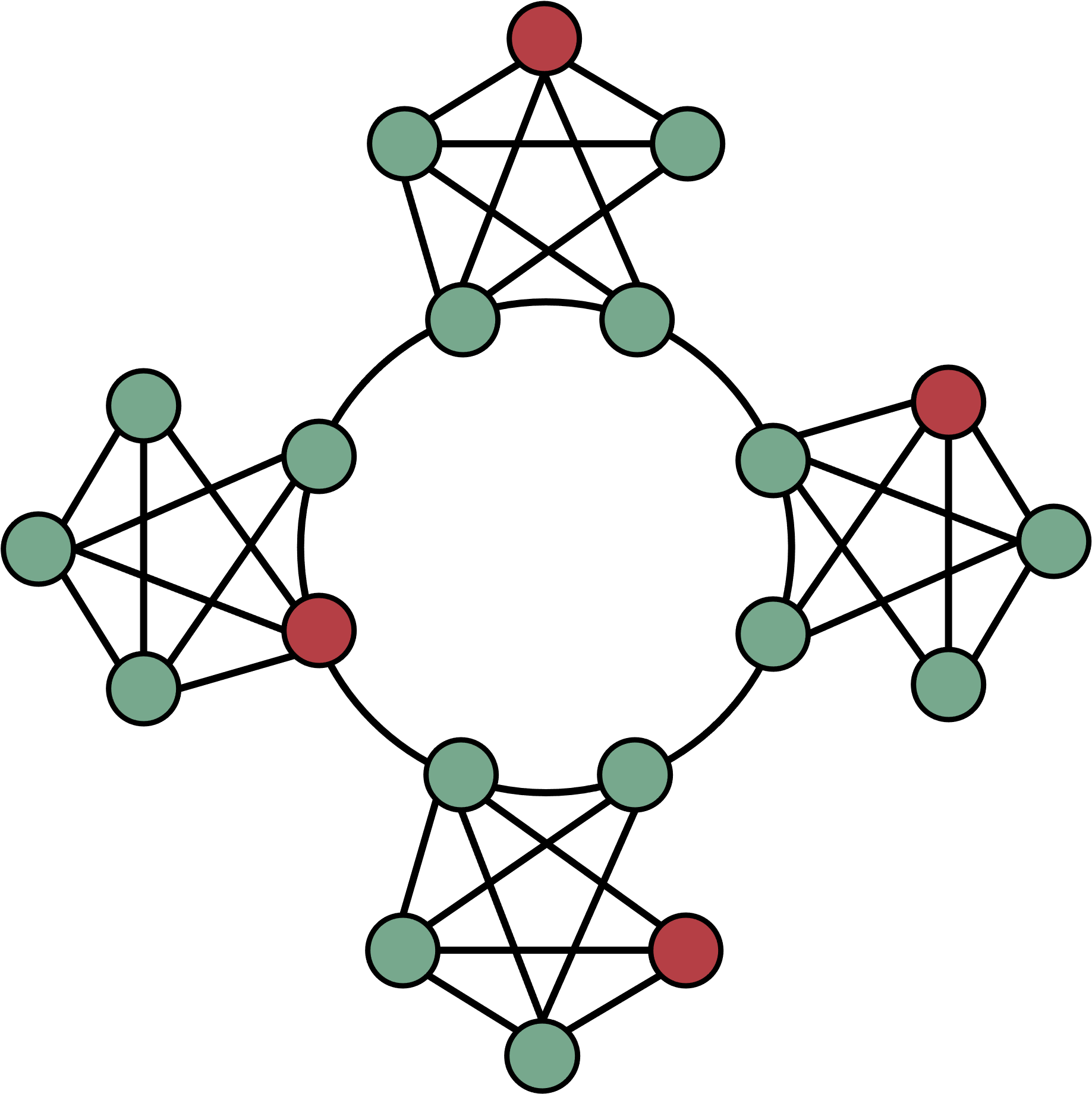}
  \caption*{Anchors of PSGNN (ours). }
\end{subfigure}
\caption{Samples of the selected anchors by the P-GNN~\cite{you2019position} (left) and the PSGNN proposed in this paper (right) for the connected Caveman graph. The anchors are highlighted in red. The anchors chose by the PSGNN distribute evenly in different communities without having symmetric anchors, whereas the ones by the P-GNN randomly scatter and induces symmetry. }
\label{fig:pgnn_psgnn_anchor}
\end{figure}

In this paper, we focus on the graphs whose node features are all identical. This setting has applications in the real world when the node properties are hard or sensitive to obtain. Under this scenario, we solely rely on the network's topological structure to conduct various downstream tasks such as community detection and link prediction. Information of pairwise distances between two nodes then becomes the indispensable only cue. Hence, we require the graph neural networks that are capable of position-awareness for nodes. 

To empower GNNs with being position-aware, recently, You~\etal~\cite{you2019position} proposed position-aware graph neural networks (P-GNNs) by setting some nodes within the graph as anchors. Thereby, other nodes can know their relative distances by comparing their distances to anchors~\cite{you2019position}. However, P-GNNs select anchors in a random fashion~\cite{you2019position}. When the anchors are ill-selected, two issues arise: 
1) Anchors are not evenly distributed within the graph. In P-GNNs, nodes learn their feature vectors via aggregating the representations of different anchors,
and anchors extract features via aggregating information of their neighbors. Thus, if the anchors concentrate in a small subgraph (see the left of \autoref{fig:pgnn_psgnn_anchor}), anchors cannot extract comprehensive information of a graph, leading to the poor representations of nodes far away from the anchors. 
2) Anchors locate on symmetric nodes. Given the condition that all the nodes have the same features, two nodes are of symmetry if their learned representations are the same via message passing and aggregating. That is, the two subgraphs around the two nodes are isomorphic. If two anchors are symmetric, the distances from two symmetric nodes to the two anchors will be the same, causing position-unawareness. 

To address the drawbacks due to random anchor selection mentioned above, we propose Position-Sensing Graph Neural Networks (PSGNNs). Instead of selecting anchors arbitrarily, the PSGNN contains a learnable anchor-selection component, enabling the network to choose anchors in a back-propagatable approach. Theoretically, we show selecting a case of the smallest set of anchors is NP-complete. This reflects the hardship of selecting anchors, leading to the impracticability of choosing optimal anchors with an efficient deterministic algorithm. 
Thus, selecting anchors through heuristics approaches like learning is inevitable. 
Practically, PSGNNs exhibit substantial improvement over a range of existing competitive GNNs on many graph datasets. Subsequent analysis discovers that PSGNNs select well-distributed and asymmetric anchors (see the right one in \autoref{fig:pgnn_psgnn_anchor}). 
Furthermore, the performance of the PSGNN steadily grows as the graph size enlarges, reflecting its desirable scalability. 

We summarize our contributions as follows: 
\begin{enumerate}
    \item We show a case of selecting the minimum number of anchors is NP-complete. 
    \item We propose PSGNNs that can learn to select ideal anchors in a back-propagatable fashion, outperforming various existing state-of-the-art models. 
    \item PSGNNs also reveal desirable scalability, learning relative positions between nodes well on giant graphs. 
\end{enumerate}

\section{Related Work}
Sperduti \etal first utilized neural networks to directed
acyclic graphs, which motivated pioneering research on GNNs. Then, a series of work on recurrent graph neural networks (RecGNNs) emerged, such as~\cite{gallicchio2010graph}. 

Encouraged by CNNs' success on computer vision, researchers start applying convolutional operations to graphs. These approaches are under the umbrella of convolutional graph neural networks, such as~\cite{bacciu2018contextual, micheli2009neural}. Spectrum-based graph convolutional networks is a subclass of convolutional graph neural networks~\cite{bianchi2020spectral}. 
Spectrum-based GNNs map input graphs to spectrum representations, indicating the degree of diffusion~\cite{bruna2013spectral}. 

Original spectrum-based GNNs directly utilize Laplacian transformation~\cite{pavez2016generalized} and the convolution theorem~\cite{xu2018graph} to conduct convolutional operations on graphs~\cite{maretic2020graph}. To reduce the computational cost of Laplacian transformation, the Chebyshev approximation has been applied to address the high computational-cost operations~\cite{defferrard2016convolutional}. Using the first two terms of the Chebyshev polynomials, Kipf~\etal\cite{kipf2016semi} proposed graph convolutional networks. Following \cite{kipf2016semi}, various other GNNs have been conceived, including both the spectral-based approaches and
the spatial-based approaches, such as graph attention networks (GATs)~\cite{velivckovic2017graph}, graph sampling $\&$ aggregate (GraphSAGE)~\cite{hamilton2017inductive}, and graph isomorphism networks (GIN)~\cite{xu2018powerful}. 
On the other hand, a range of network embeddings methods have also been presented, such as DeepWalk~\cite{perozzi2014deepwalk}, LINE~\cite{tang2015line}, and node2vec~\cite{grover2016node2vec}. In a recent advancement, Peng~\etal\cite{peng2022reverse} proposed a reverse graph learning approach for GNNs, addressing the initial graph's imperfections and extending GNN applications to out-of-sample data points.

All the mentioned approaches follow the message passing and aggregating paradigm. GATs assign different weights to the messages received from distinct neighbors~\cite{velivckovic2017graph}. For GraphSAGE, a node receives messages from a sampled set of neighbors~\cite{hamilton2017inductive}, instead of collecting embeddings from all the neighbors, and graph isomorphism networks (GINs) are inspired by the graph isomorphism problem to build GNNs that have the equivalent capability with the Weisfeiler-Lehman (WL) test~\cite{xu2018powerful}.

Despite the improved performance, GNNs based on message passing and aggregating suffer from the over-smoothing problem~\cite{chen2020measuring} \ie as the number of network layers increases, the performance of GNNs decreases. To address over-smoothing, Li~\etal~\cite{li2019deepgcns} proposed deep graph convolutional networks using residual connections to improve the efficiency of back-propagation. 

To study the theoretical power of GNNs, Xu~\etal~\cite{xu2018powerful} state that the capability of existing GNNs in differentiating two isomorphic graphs does not exceed the 1-Weisfeiler-Lehman (1-WL) test~\cite{shervashidze2011weisfeiler}. In other words, without assigning different nodes with distinct features, many existing GNNs, \eg GCNs~\cite{kipf2016semi} and GraphSage~\cite{hamilton2017inductive}, embed nodes in symmetric locations into the same embeddings. 

Hence, although the above GNNs demonstrate excellent performance in some tasks like node classification, the node embeddings learned by these GNNs do not contain relative positions for two nodes in symmetric locations. To enable position-awareness in GNNs, You~\etal~\cite{you2019position} presented position-aware graph neural networks (P-GNNs) and are the most relevant model to our proposed model. P-GNNs set some nodes within the graph as anchors so that other nodes can know their relative distances from the designated anchors. Building on the framework of~\cite{you2019position}, Huo~\etal\cite{huo2023trustgnn} and Tiezzi~\etal\cite{tiezzi2022graph} further explore GNNs in trust evaluation and graph drawing, respectively, demonstrating the versatility and impact of P-GNNs.

You~\etal~\cite{you2019position}  also introduced the identity-aware GNN (ID-GNN) that can extend the expressing capability of GNNs beyond the 1-WL test. However, the work focuses on empowering different nodes with distinct representations instead of making GNNs position-aware. That is, the embeddings of two nodes learned by ID-GNNs do not preserve their distance information, whereas this information is retained with low distortion by P-GNNs.  

Regardless of the possibility of learning relative distances between graph nodes, P-GNNs select anchors randomly.  This paper shows that the random selection of anchors leads to performance compromise and position unawareness. Moreover, we present a solution model: Position-Sensing Graph Neural Networks, to address the problems of random anchor selection. 
\section{Preliminaries}
\subsection{Graph Neural Networks}
Graph neural networks (GNNs) are able to learn the lower-dimensional embeddings of nodes or graphs~\cite{wu2020comprehensive}. In general, GNNs apply a neighborhood aggregation strategy~\cite{wu2020comprehensive} iteratively. The strategy consists of two steps: 
\begin{gather}
    a_v^{(l)} = \text{Aggregate}^{(l)} (\{ h_u^{(l-1)} |  u \in \text{Neighbor}(v) \}), \notag \\
    h_v^{(l)} = \text{Combine}^{(l)} (h_v^{(l-1)}, a_v). \notag 
\end{gather}
In the above equations, $\text{Aggregate}(\cdot)$ is the function to aggregate all the information of the neighbor nodes around node $v$. Moreover, $h_v^{(l)}$ stands for the embedding of node $v$ at layer $l$. The iterations continue for $l$ steps. The number $l$ is the layer number. To reduce verbosity, we refer to the GNNs following the mentioned pattern of aggregating and combining simply as standard GNNs. 

\begin{figure}[t]
\centering
\begin{subfigure}{.2\textwidth}
  \centering
  \includegraphics[width=.99\linewidth]{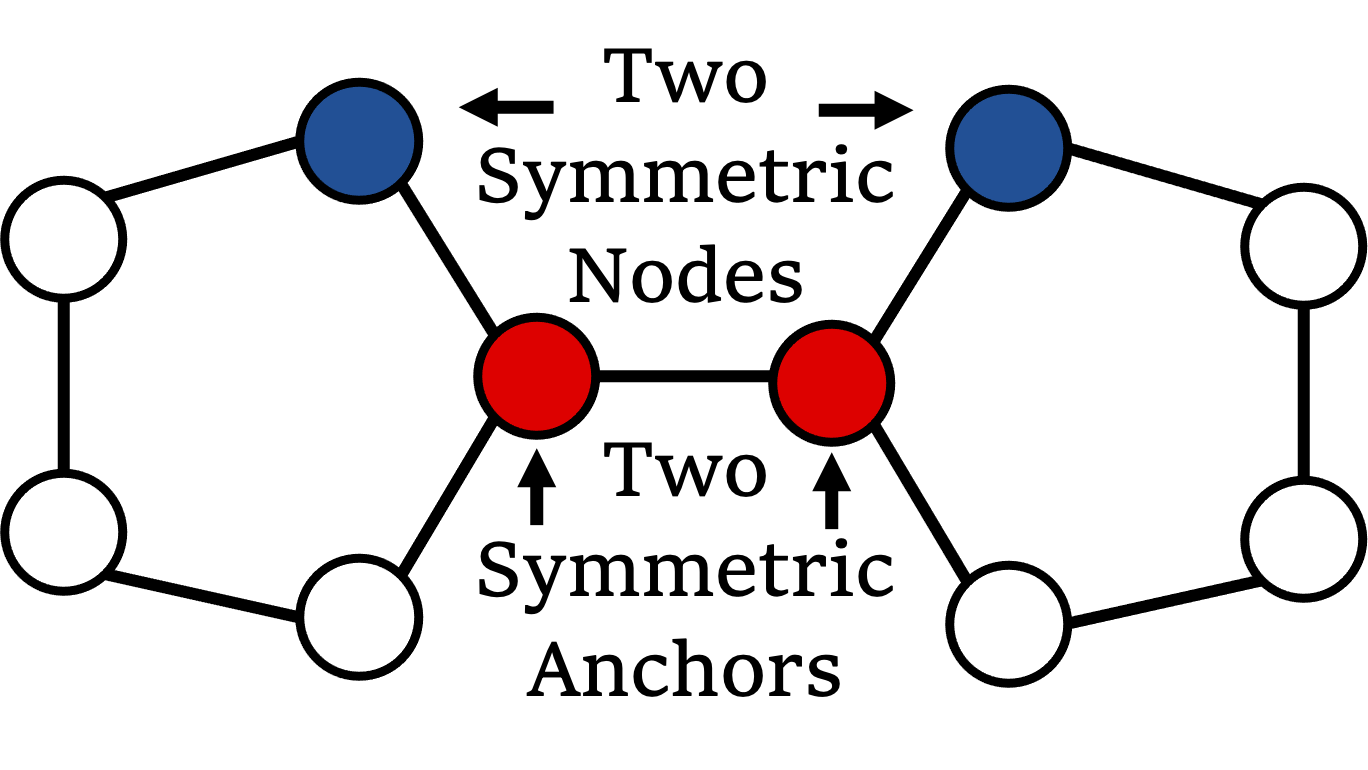}  
  \caption{Without anchors }
  \label{fig:sub-first}
\end{subfigure}
\hspace{0.3cm}
\begin{subfigure}{.2\textwidth}
  \centering
  \includegraphics[width=.99\linewidth]{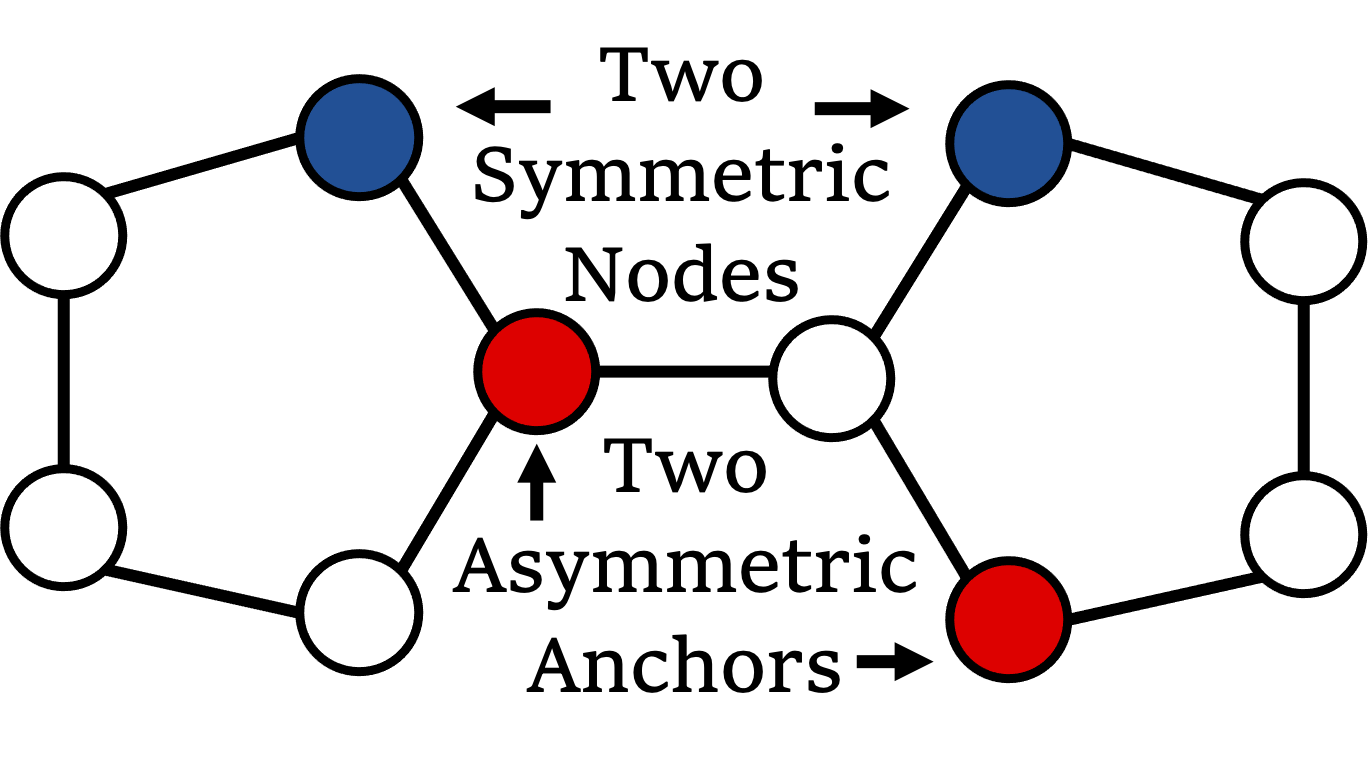}  
  \caption{With an anchor }
  \label{fig:sub-second}
\end{subfigure}
\caption{Illustration of symmetric node positions. The two blue nodes are in symmetric locations within each graph. The dotted blue node represents an anchor. }
\label{fig:sym_nodes_wt_no_anchor}
\end{figure}

\begin{figure}[t]
\centering
\begin{subfigure}{.24\textwidth}
  \centering
  \includegraphics[width=.85\linewidth]{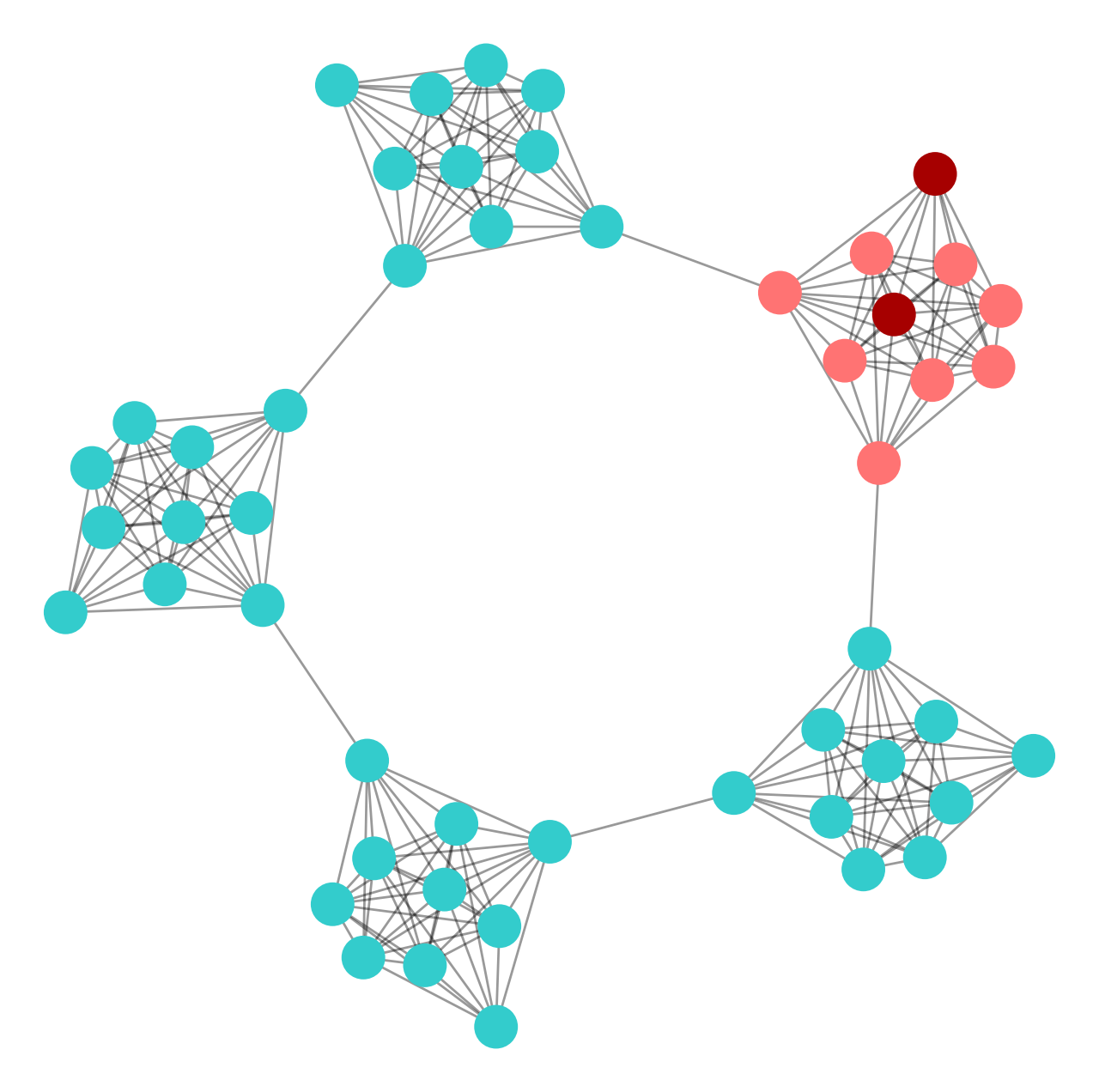}  
  \caption{Non-Overlapping Coverage}
  \label{fig:sub-first}
\end{subfigure}
\begin{subfigure}{.24\textwidth}
  \centering
  \includegraphics[width=.85\linewidth]{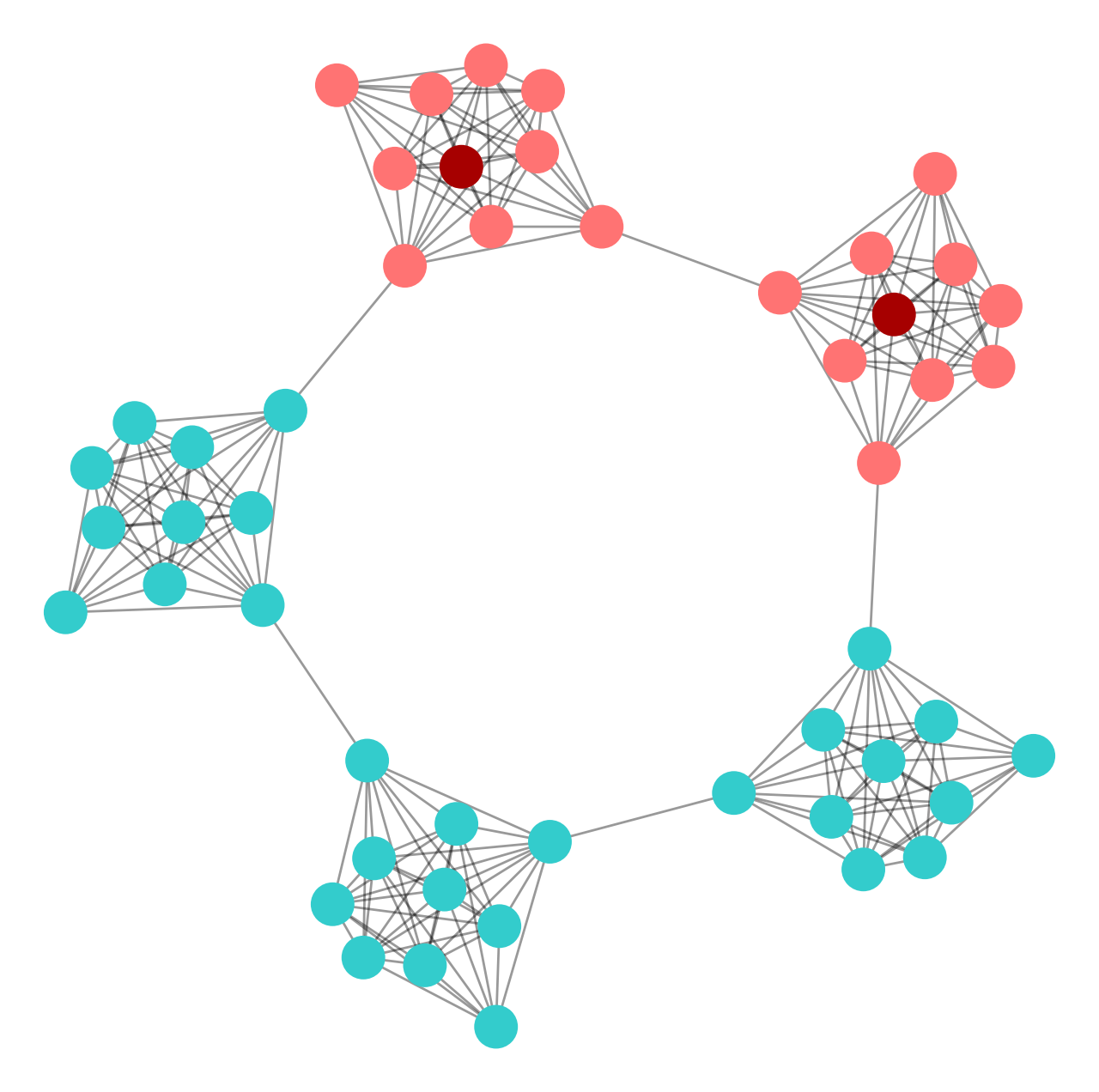}
  \caption{Overlapping Coverage}
  \label{fig:sub-second}
\end{subfigure}
\caption{Illustration of the covered nodes (yellow nodes) of anchors (red nodes).  }
\label{fig:anchor_coverage}
\end{figure}

\section{Relative Positions Learning}
\label{sec:learn_via_anchors}
To reiterate, P-GNNs~\cite{you2019position}
aim to enable position-awareness during GNNs learning embeddings via message passing and aggregating.  
For example, as \autoref{fig:sym_nodes_wt_no_anchor} illustrates, the two symmetric nodes will be assigned with the same embeddings by a standard GNN, but their distances to the anchors are different~\cite{you2019position}. Therefore, using the distances from every node to the anchors, P-GNNs will know the relative positions among nodes.

Although P-GNNs have been equipped with position-awareness of nodes, P-GNNs select anchor sets in a random fashion. Such randomness leads to two problems. 

First, the neighbors of randomly selected anchors may not cover all the nodes within a graph. Here, we say if a node is a $d$-hop neighbor of an anchor, where $d$ is the number of layers for a GNN to learn anchor embeddings, we then say the anchor covers the node. We consider that anchors are better if they can cover every node within the graph because we assume such anchors together contain more knowledge about the features and structures of the graph than anchors covering only a subset of graph nodes. For example, \autoref{fig:anchor_coverage} exhibits the anchors' cases covering more and fewer nodes. 
We consider the coverage of anchors being non-overlapped is better than being overlapped since the mutually non-overlapping anchor coverage leads to different anchors containing information of distinct nodes. Thus, the features of two anchors are more likely to be orthogonal. 

Second, as Liu~\etal~\cite{liu2020graph} have also pointed out; if anchors locate on symmetric nodes, symmetric nodes will have the same distances to the anchors. Consequently, P-GNNs become position-unaware for some graph nodes. For example, as \autoref{fig:sub-first} illustrates, 
two anchors (the red nodes) are located on symmetric locations. As a result, the two symmetric nodes will then have the same embeddings learned from a P-GNN since the two distances to the two anchors from the two nodes are the same. The symmetric anchors then cause the embeddings of the two nodes not to be differentiable, making P-GNNs fail to capture the relative positions between them.

Our proposed approach is motivated by these two limitations. In sum, both problems root in flawed anchor selection. One intuitive idea can be to devise deterministic algorithms for choosing anchors. However, in the following, we show that selecting anchors is not a trivial problem. Concretely, we define the set of anchors of a graph $g=(\vertex, \edge)$ as a set of nodes $\mathcal{A} \subseteq \vertex$. We define that anchor $a$ covers node $v$ if the distance between them is less than or equal to $d$, where $d$ is a parameter to be set beforehand. We see that when $d = 1$, selecting the smallest set of anchors that cover the entire graph is equivalent to the smallest dominant set problem. The latter one is NP-complete, reflecting the hardship of the anchor-selection problem.

\begin{figure*}
    \centering
    \includegraphics[width=0.9\linewidth]{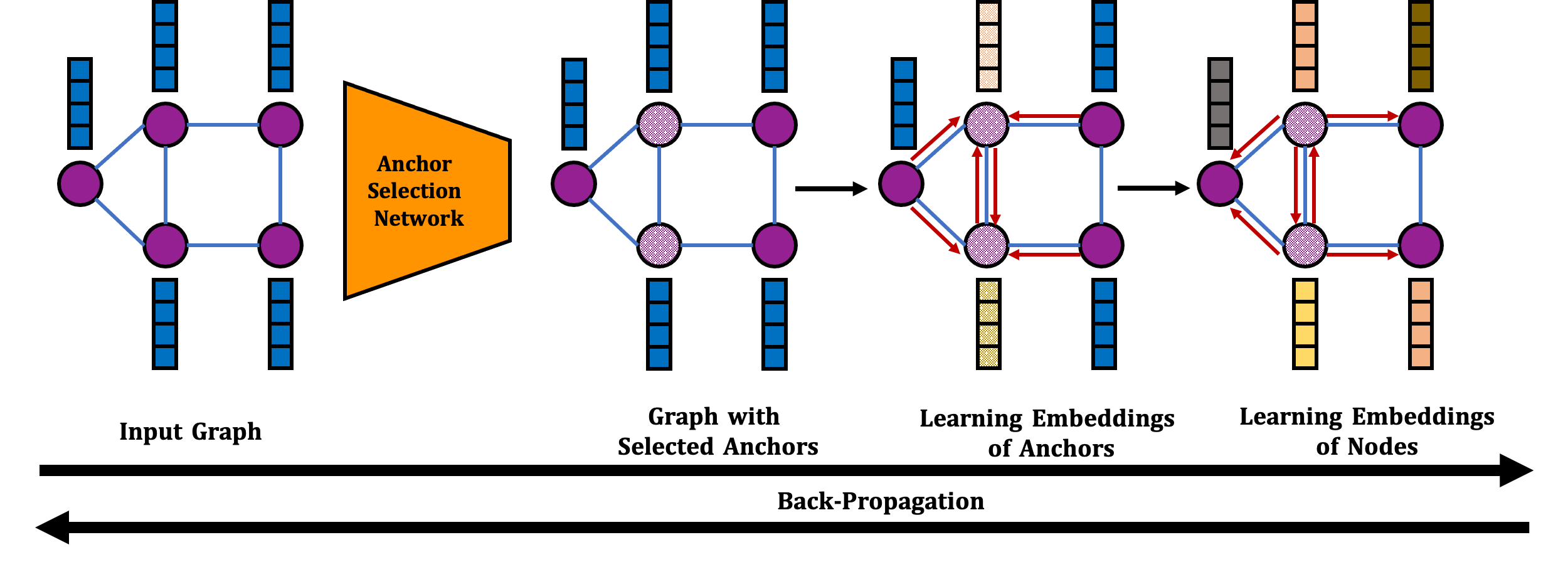}
    \caption{Training process of a PSGNN. Dotted notes represent anchors. }
    \label{fig:my_label}
\end{figure*}

\section{Our Approaches}
\RestyleAlgo{boxruled}

\RestyleAlgo{boxruled}
\begin{algorithm}[h!]
\caption{Procedures of the PSGNN for an epoch. }
\SetAlgoLined
\KwInput{Graph $G = (\mathcal{V}, \mathcal{E})$; Node input features $\{ \x_v^{(0)} \}$}
\KwOutput{Position-Aware Node embeddings }
--------------------------------------------------------------------
\\
\tcp{(1) anchor selection component. }
\For {$l = 1, ..., D$} {
\tcp{synthesizing features for selecting anchors. }
        \For {$v \in \mathcal{V}$} {
        \tcp{collect neighbor features. }
            $\mathbf{x}_v^{(l)} = 
            \gamma_\phi \left( \mathbf{x}_v^{(l-1)}, \msquare_{j \in \mathcal{N}(i)} \big(\mathbf{x}_i^{(l-1)}, \mathbf{x}_j^{(l-1)},\mathbf{e}_{j,i}\big) \right)
            $\;
        }
    }
    \tcp{add noise for exploration. }
    $\mathbf{o} = n \big( f_\phi^{(1)}(\mathbf{x}^{(D)}) \big) + \alpha \cdot \epsilon \sim \mathcal{N}(\mathbf{0}, \mathbf{I})$ \;
    \tcp{select $K$ anchors, obtain their embeddings. }
    $\zeta = \argsort (\mathbf{o}_1, ..., \mathbf{o}_V)$ \; 
        $\mathbf{m} = \bool( \zeta \le K$) \; 
        $\mathcal{A} = \select( \mathbf{x}^{(D)}, \mathbf{m} ) \in \mathbb{R}^{K \times R}$ \;
--------------------------------------------------------------------
\\
\tcp{(2) positional feature learning component. }
\For {$l = 1, ..., L$}{
        \For {$v \in \mathcal{V}$}{
            $\mathbf{M}_v = \mathbf{0} \in \mathbb{R}^{K \times R}$ \; 
            \For{$i = 1, ..., K$}{
                \tcp{calculate the distance between node $v$ and anchor $\mathcal{A}_{i}$. }
                $\mathbf{d} = d (v,  \mathcal{A}_{i})$ \; 
                \tcp{transform the distance to an embedding.}
                $\mathbf{h}_{\mathbf{d}} = f_\phi^{(2)} (\mathbf{d})$ \; 
                \tcp{concatenate the distance embedding with the node embedding. }
                $\mathbf{M}_v[i] \leftarrow \CAT(\mathbf{h}_{\mathbf{d}}, \mathbf{h}_v)$ \; 
                \tcp{applying an MLP on the concatenated representations. }
                $\mathbf{M}_v[i] \leftarrow f_\phi^{(3)}(\mathbf{M}_v[i])$ \;
            }
            \tcp{aggregate representations of every anchor in tensor $\mathcal{A}$ over node $v$. }
            $\mathbf{h}_v = \AGG( \{ 
                \mathbf{M}_v[i], \forall i \in [1,K]
            \} )$ \;
        }
    }
\label{alg:ps_gnn}
\end{algorithm}

\subsection{Problem Formulation}
We aim to create a neural network that learns to pick the representative anchors in order to conduct various downstream tasks, where anchors are previously defined as a set of nodes $\mathcal{A} \subseteq \vertex$. 
Features of these anchors comprise information of the entire graph. Mathematically, we can express model $s_{\phi}$ for selecting anchors as:
\begin{align}
    s_{\phi}: (\vertex, \edge) \rightarrow \{ v_i \}_{i=1}^{K}, 
    \label{eq:anchor_pick}
\end{align}
where $\phi$ denotes the parameters of model $s$, $\{ (\vertex, \edge)_i \}_{i=1}^{N}$ is a set of graphs, $\vertex$ and $\edge$ correspond to vertices and edges, and $\{ v_i \}_{i=1}^{K}$ is the picked $K$ anchors. 
Again, we define a set of anchors to be a subset of vertices within a graph. 
We consider a set of anchors to be optimum if for every vertex, its distance to the nearest anchor within the anchor set is smaller than $\mathcal{D}$, where $\mathcal{D}$ is a hyperparameter to be set. Formally, given a predefined value $\mathcal{D}$, anchor set $\mathcal{A}$ is optimum if: 
\begin{equation}
    \forall v \in \mathcal{V}, \exists a \in \mathcal{A}: d(a, v) \le \mathcal{D}, 
\end{equation}
where function $d$ evaluates the distance between two nodes within a graph. 

\subsection{Position-Sensing Graph Neural Network}
We desire to devise a learning model that can achieve the anchor selection process of \autoref{eq:anchor_pick} in a back-propagatable manner, followed by certain downstream tasks. To this end, we design the position-sensing graph neural network (PSGNN). It comprises of two components: 1) the anchor selection component and 2) the positional feature learning component. In the following, we describe the two components in detail.

\subsection{Anchor Selection Component}
The anchor selection component is a composition of a feature aggregator $\mathcal{F}_\phi$ and an anchor picker $\mathcal{P}_\phi$, returning a vector $\mathbf{o}$, expressed as: 
\begin{align}
    \mathcal{P}_\phi \circ \mathcal{F}_\phi \big[ (\vertex, \edge) \big] = \mathbf{o}. 
\end{align}
The algorithmic procedure is outlined in the first part of \autoref{alg:ps_gnn}. 
In the following, we concretely describe feature aggregator $\mathcal{F}_\phi$ and anchor picker $\mathcal{P}_\phi$. 

\textbf{Feature Aggregator}: 
The feature aggregator component $\mathcal{F}_\phi$ can be an arbitrary GNN model such as GCN~\cite{kipf2016semi}, GAT~\cite{velivckovic2017graph}, and GIN~\cite{xu2018powerful}. 
For each layer's operation of these models can be formulated as: 
\begin{equation}
    \mathbf{x}_v^{(l)} = 
            \gamma_\phi \left( \mathbf{x}_v^{(l-1)}, \msquare_{j \in \mathcal{N}(i)} \big(\mathbf{x}_i^{(l-1)}, \mathbf{x}_j^{(l-1)},\mathbf{e}_{j,i}\big) \right), 
\end{equation}
where
$\msquare$ stands for the operation of passing messages and aggregating information of neighboring nodes, 
$l$ represents the layer, 
$\mathbf{e}_{j,i}$ denotes feature of the edge between nodes $j$ and $i$,\footnote{We include the notion of edge features here for completeness. In our implementation, no edge feature is utilized. }  
$\mathcal{N}(i)$ stands for the set of neighbors of node $i$, 
and $\gamma_\phi$ is a function for updating the node's feature based on its own embedding and the aggregated neighboring information. 

\textbf{Anchor Picker}: 
The key idea for anchor picker $\mathcal{P}_\phi$ is to transform each node's embedding to a scalar, indicating the likelihood of the node being selected as an anchor. The anchor picker comprises of four steps to achieve this end. 

\textit{(1) Transforming Features}: First, we apply an affine layer to transform the node embeddings to be scalar values, followed by a normalization layer. In our implementation, the normalization is $\ell_2$-norm. Other normalization layers can also be used. 
We denote this process as 
$f_\phi^{(1)}(\mathbf{x}^{(D)})$, 
where $\mathbf{x}^{(D)}$ stands for the node embeddings from feature aggregator $\mathcal{F}_\phi$. 

\textit{(2) Adding Noise}: Next, we need to trade-off between exploration and exploitation of selecting anchors. Exploration is to select the anchors that have not been selected before, and exploitation is to stick using the previously-selected anchors that lead to low loss values. To this end, we add a randomly generated noise $\epsilon \sim \mathcal{N}(\mathbf{0}, \mathbf{I}) \in \mathbb{R}^N$ to the likelihood vector $\mathbf{o}$: 
\begin{equation}
    \mathbf{o} + \alpha \cdot \epsilon, 
    \label{eq:o_epsilon}
\end{equation}
where $\alpha$ is a hyperparameter controlling the trade-off. Larger $\alpha$ indicates more randomness. That is, it is to improve the searching space. We set $\alpha$ based on the performance evaluated on the validation set. 

\textit{(3) Creating the Boolean Mask}: Up to here, every node has a value indicating its likelihood of being selected as an anchor. We use $\mathbf{o} \in \mathbb{R}^{N} = <\mathbf{o}_1, ..., \mathbf{o}_N>$ to denote these values, where each entry corresponds to a node's likelihood. We then desire to select embeddings of $K$ nodes whose likelihoods are the top-$K$ largest. We implement this desirability as follows to enable differentiability. First, we perform $\argsort$ on these $N$ values, where the larger values result in smaller indices. Then, we create a boolean mask $\mathbf{m}$ such that the indices larger than $K$ will have mask value $0$ and the others have mask value $1$. That is, the nodes that are the top-$K$ most likely to be anchors will have the mask value $1$.

\textit{(4) Selecting Anchors}: Finally, using the above boolean mask, we select embeddings of the nodes corresponding to the top-$K$ largest likelihoods. 
They form the anchor tensor where each row corresponds to an anchor's embedding, denoted as $\mathcal{A} \in \mathbb{R}^{K \times R}$. 

To illustrate, if we are to choose $K = 2$ anchors, then given the assumption where after adding noise $\epsilon$, the $2$-nd and $4$-th dimensions are of the top-$2$ maximum within the output vector~$\mathbf{o}$ of the anchor selection component, the nodes with IDs $2$ and $4$ will be selected as the anchors. Then, we extract their embeddings as the anchor tensor for the subsequent feature learning. 

\subsection{Positional Feature Learning Component} 

Once we have determined the anchors, we then launch the positional feature learning component to extract the position-aware node embeddings. The detailed algorithmic steps are outlined in the second part of \autoref{alg:ps_gnn}. To be more specific, for each node $v$, we iterate every anchor $\mathcal{A}_i$ within the anchor tensor $\mathcal{A}$ to assign an anchor for node $v$, where $i$ stands for an index of the tensor. We initialize a tensor $\mathbf{M}_v$, where each dimension records the representation of an anchor over node $v$. For example, $\mathbf{M}_v[i]$ stands for the information of anchor $i$ on node $v$. This information is obtained with the following steps: 
\begin{enumerate}
    \item For each anchor $\mathcal{A}_i$ within the anchor tensor $\mathcal{A}$, we calculate its distance to the node $v$ and map the distance to an embedding: $d(v, \mathcal{A}_i)$.\footnote{In practical implementation, calculating the shortest distances can be performed as a preprocessing step. We can then store these pre-calculated distances for future access to reduce computational complexity. } 
    \item We increase the dimension of the distance's representations as $\mathbf{h}_{\mathbf{d}}$ with an MLP: $f_\phi^{(2)}(\mathbf{h}_{\mathbf{d}})$. 
    \item We concatenate the representations of the distance with the node's embedding, recorded in $\mathbf{M}_v[i]$: $\mathbf{M}_v[i] \leftarrow \CAT(\mathbf{h}_{\mathbf{d}}, \mathbf{h}_{v})$. 
    \item We apply an MLP on the concatenated feature: $\mathbf{M}_v[i] \leftarrow f_\phi^{(3)}(\mathbf{M}_v[i])$.
    \item Once we have obtained an feature of every anchor for node $v$, we apply an aggregating operators on these $K$ features: $\mathbf{h}_v = \AGG( \{ 
                \mathbf{M}_v[i], \forall i \in [1,K]
            \} )$. Concretely, we map every row's representation to a scalar, followed by a reshape operator. Then, every node's reprentation will be of dimension $K$, where each entry corresponds to an anchor.  
\end{enumerate}

The most consuming component of the proposed algorithm lies on calculating the complexity of pairwise note distances. For example, the time complexity is $O(|\mathcal{V}|^3)$
and space complexity is $O(|\mathcal{V}|^2)$ if using the Floyd–Warshall algorithm. Nonetheless, calculating pairwise node distances can happen in a preprocessing step before the neural network operates. If we can access these distances in $O(1)$, the time and space complexity of the proposed PSGNN for one layer are $O(K \cdot \mathcal{V})$ and $O(1)$ respectively.

\subsection{Working Mechanisms}
Our PSGNN learns to select anchors in a trial-and-error approach. With back-propagation, it can choose increasingly more appropriate anchors for downstream tasks. We show the intuitive working mechanisms here.

Suppose there is only one anchor, and the node with ID~3 (denote as node~3 for brevity) is the ideal anchor. However, the anchor-selection component wrongly chooses node~4 as the anchor. That is, the likelihood in the 4-th entry of output vector $\mathbf{o}$ is larger than that of the 3-th entry, \ie{} $\mathbf{o}_4 > \mathbf{o}_3$. 
Since PSGNN contains randomness in selecting anchors, given enough times of sampling, node~3 will be selected as an anchor at some time. As node~3 is more appropriate to be an anchor than node~4, the loss value is likely to be smaller. As a result, back-propagation will increase $\mathbf{o}_3$. Next time, $\mathbf{o}_3$ will have higher chance in selecting node~3. 
\begin{table}[t]
\centering
\caption{AUC results of comparing PSGNNs against existing competitive baselines on various datasets for pairwise node classification and link prediction tasks. Best results are highlighted with bold face and the red color. }
\begin{subtable}[t]{0.48\textwidth}
\caption{Pairwise node classification. } 
\resizebox{0.99\textwidth}{!}{
\begin{tabular}{lccc}
\toprule 
 & Communities & Email & Protein \\ 
\midrule 
GCN~\cite{kipf2016semi} & $0.515 \pm 0.026$ & $0.505 \pm 0.011$ & $0.506 \pm 0.003$ \\
GraphSAGE~\cite{hamilton2017inductive} & $0.525 \pm 0.011$ & $0.525 \pm 0.089$ & $0.515 \pm 0.011$ \\ 
GAT~\cite{velivckovic2017graph} & $0.608 \pm 0.014$ & $0.516 \pm 0.012$ & $0.519 \pm 0.012$ \\
GIN~\cite{xu2018powerful} & $0.617 \pm 0.101$ & $0.529 \pm 0.025$ & $0.525 \pm 0.011$ \\
\midrule  
\multicolumn{4}{c}{Position-Aware Methods} \\
\midrule 
PGNN~\cite{you2019position} & $0.912 \pm 0.071$ & $0.563 \pm 0.012$ & $0.502 \pm 0.023$ \\ 
A-GNN~\cite{liu2020graph} & $0.945 \pm 0.068$ & $0.537 \pm 0.031$ & $0.511 \pm 0.037$ \\ 
PSGNN (Ours) & \color{red}{$\mathbf{0.978 \pm 0.035}$} & \color{red}{$\mathbf{0.736 \pm 0.017}$} & \color{red}{$\mathbf{0.533 \pm 0.018}$} \\
\bottomrule 
\end{tabular}
}
\end{subtable}

\centering
\vspace{2mm}
\begin{subtable}[t]{0.48\textwidth}
\caption{Link prediction. } 
\resizebox{0.99\textwidth}{!}{
\begin{tabular}{lccc}
\toprule 
 & Communities & Grids & Bio-Modules \\ 
\midrule
GCN~\cite{kipf2016semi} & $0.435 \pm 0.037$ & $0.519 \pm 0.015$ & $0.528 \pm 0.015$ \\
GraphSAGE~\cite{hamilton2017inductive} & $0.511 \pm 0.011$ & $0.511 \pm 0.054$ & $0.531 \pm 0.008$ \\ 
GAT~\cite{velivckovic2017graph} & $0.576 \pm 0.023$ & $0.526 \pm 0.037$ & $0.544 \pm 0.007$ \\
GIN~\cite{xu2018powerful} & $0.519 \pm 0.035$ & $0.558 \pm 0.026$ & $0.511 \pm 0.017$ \\
\midrule  
\multicolumn{4}{c}{Position-Aware Methods} \\
\midrule 
PGNN~\cite{you2019position} & $0.927 \pm 0.023$ & $0.765 \pm 0.062$ & $0.913 \pm 0.015$ \\ 
A-GNN~\cite{liu2020graph} & $0.936 \pm 0.028$ & $0.812 \pm 0.083$ & $0.934 \pm 0.012$ \\ 
PSGNN (Ours) & \color{red}{$\mathbf{0.954 \pm 0.032}$} & \color{red}{$\mathbf{0.942 \pm 0.044}$} & \color{red}{$\mathbf{0.989 \pm 0.001}$} \\
\bottomrule 
\end{tabular}
}
\end{subtable}
\label{table:compare_with_sota}
\end{table}

\begin{table*}[tb]
\caption{
Scalability analysis of PSGNNs against existing competitive GNNs. Standard deviations are given.
Best results are highlighted with bold face and the red color.  }
\centering
\begin{subtable}[t]{1.0\textwidth}
\caption{AUC scores for pairwise node classification on communities with different numbers and sizes. 
C stands for the number of communities and S represents the community size.
}
\resizebox{1.0\textwidth}{!}{
\begin{tabular}{l c c c c c c c c c}
\toprule
Methods & C2; S8 & C2; S32 & C2; S64 & C32; S8 & C32; S32 & C32; S64 & C64; S8 & C64; S32 & C64; S64 \\ 
\midrule
GCN~\cite{kipf2016semi} & $0.482 \pm 0.099$ & $0.502 \pm 0.043$ & $0.556 \pm 0.029$ & $0.518 \pm 0.033$ & $0.525 \pm 0.024$ & $0.534 \pm 0.004$ & $0.510 \pm 0.016$ & $0.532 \pm 0.004$ & $0.534 \pm 0.016$ \\
GraphSAGE~\cite{hamilton2017inductive} & $0.590 \pm 0.133$ & $0.500 \pm 0.014$ & $0.510 \pm 0.005$ & $0.498 \pm 0.020$ & $0.500 \pm 0.004$ & $0.507 \pm 0.000$ & $0.487 \pm 0.035$ & $0.500 \pm 0.004$ & $0.502 \pm 0.011$ \\
GAT~\cite{velivckovic2017graph} & $0.520 \pm 0.108$ & $0.500 \pm 0.027$ & $0.562 \pm 0.022$ & $0.510 \pm 0.024$ & $0.534 \pm 0.002$ & $0.519 \pm 0.011$ & $0.487 \pm 0.020$ & $0.515 \pm 0.025$ & $0.512 \pm 0.004$ \\
GIN~\cite{xu2018powerful} & $0.432 \pm 0.203$ & $0.505 \pm 0.024$ & $0.507 \pm 0.016$ & $0.472 \pm 0.021$ & $0.500 \pm 0.005$ & $0.518 \pm 0.015$ & $0.468 \pm 0.039$ & $0.509 \pm 0.014$ & $0.518 \pm 0.016$ \\

\midrule 
\multicolumn{10}{c}{Position-Aware Methods} \\
\midrule 
P-GNN~\cite{you2019position} & $0.756 \pm 0.172$ & $0.773 \pm 0.041$ & $0.770 \pm 0.063$ & $0.875 \pm 0.121$ & $0.912 \pm 0.074$ & $0.856 \pm 0.113$ & $0.882 \pm 0.071$ & $0.925 \pm 0.078$ & $0.892 \pm 0.093$ \\
A-GNN~\cite{liu2020graph} & $0.723 \pm 0.166$ & $0.826 \pm 0.014$ & $0.831 \pm 0.014$ & $0.980 \pm 0.018$ & $0.979 \pm 0.022$ & $0.983 \pm 0.001$ & $0.972 \pm 0.014$ & $0.961 \pm 0.024$ & $0.972 \pm 0.001$ \\
PSGNN & \color{red}{$\mathbf{0.842 \pm 0.081}$} & \color{red}{$\mathbf{0.843 \pm 0.024}$} & \color{red}{$\mathbf{0.857 \pm 0.002}$} & \color{red}{$\mathbf{0.981 \pm 0.007}$} & \color{red}{$\mathbf{0.983 \pm 0.008}$} & \color{red}{$\mathbf{0.992 \pm 0.003}$} & \color{red}{$\mathbf{0.994 \pm 0.004}$} & \color{red}{$\mathbf{0.995 \pm 0.003}$} & \color{red}{$\mathbf{0.997 \pm 0.007}$} \\
\bottomrule
\end{tabular}
}
\end{subtable}
\label{tab:node_cls_com}

\begin{subtable}[t]{1.0\textwidth}
\caption{AUC scores for link prediction of the grid dataset with various sizes. Standard deviations are given. Bold texts represent the best performance. The splitting line in the middle divides position-aware methods from position-unaware methods. }
\resizebox{1.0\textwidth}{!}{
\begin{tabular}{l c c c c c c c c c}
\toprule
Methods & $16 \times 16$ & $20 \times 20$ & $24 \times 24$ & $28 \times 28$ & $32 \times 32$ & $36 \times 36$ & $40 \times 40$ & $44 \times 44$ & $48 \times 48$ \\ 
\midrule
GCN~\cite{kipf2016semi} & $0.490 \pm 0.005$ & $0.527 \pm 0.022$ & $0.496 \pm 0.010$ & $0.498 \pm 0.011$ & $0.521 \pm 0.001$ & $0.494 \pm 0.030$ & $0.507 \pm 0.011$ & $0.517 \pm 0.016$ & $0.516 \pm 0.022$ \\
GraphSAGE~\cite{hamilton2017inductive} & $0.560 \pm 0.043$ & $0.587 \pm 0.069$ & $0.616 \pm 0.070$ & $0.567 \pm 0.030$ & $0.615 \pm 0.024$ & $0.647 \pm 0.013$ & $0.685 \pm 0.042$ & $0.636 \pm 0.043$ & $0.647 \pm 0.018$ \\
GAT~\cite{velivckovic2017graph} & $0.616 \pm 0.065$ & $0.604 \pm 0.056$ & $0.647 \pm 0.011$ & $0.615 \pm 0.001$ & $0.655 \pm 0.022$ & $0.621 \pm 0.028$ & $0.591 \pm 0.022$ & $0.586 \pm 0.042$ & $0.636 \pm 0.030$ \\
GIN~\cite{xu2018powerful} & $0.464 \pm 0.009$ & $0.509 \pm 0.028$ & $0.418 \pm 0.022$ & $0.458 \pm 0.020$ & $0.482 \pm 0.001$ & $0.459 \pm 0.020$ & $0.426 \pm 0.013$ & $0.466 \pm 0.017$ & $0.452 \pm 0.003$ \\

\midrule 
\multicolumn{10}{c}{Position-Aware Methods} \\
\midrule 
P-GNN~\cite{you2019position} & $0.653 \pm 0.162$ & $0.628 \pm 0.072$ & $0.643 \pm 0.054$ & $0.526 \pm 0.022$ & $0.591 \pm 0.038$ & $0.623 \pm 0.065$ & $0.657 \pm 0.089$ & $0.591 \pm 0.023$ & $0.565 \pm 0.040$ \\
A-GNN~\cite{liu2020graph} & $0.853 \pm 0.071$ & $0.897 \pm 0.013$ & $0.903 \pm 0.034$ & $0.881 \pm 0.067$ & $0.918 \pm 0.063$ & $0.953 \pm 0.019$ & $0.933 \pm 0.023$ & $0.951 \pm 0.021$ & $0.934 \pm 0.053$ \\
PSGNN & \color{red}{$\mathbf{0.901 \pm 0.023}$} & \color{red}{$\mathbf{0.924 \pm 0.021}$} & \color{red}{$\mathbf{0.945 \pm 0.019}$} & \color{red}{$\mathbf{0.949 \pm 0.024}$} & \color{red}{$\mathbf{0.952 \pm 0.013}$} & \color{red}{$\mathbf{0.955 \pm 0.003}$} & \color{red}{$\mathbf{0.956 \pm 0.032}$} & \color{red}{$\mathbf{0.961 \pm 0.008}$} & \color{red}{$\mathbf{0.963 \pm 0.007}$} \\
\bottomrule
\end{tabular}
}
\end{subtable}
\label{tab:size_analysis}
\end{table*}

\begin{table}[t]
    \caption{The complexity of selecting an anchor with rule-based heuristics and learning methods, as well as their AUC results on communities for pairwise node classification (Pair Node Cls) and on grids for link prediction (Link Pred). 
    The complexity is just for determining the anchors (the anchor selection component). 
    The best results are highlighted in bold face with the red color. }
    \centering
    \resizebox{0.45\textwidth}{!}{
    \begin{tabular}{l c | c c}
    \toprule 
    & & \multicolumn{2}{c}{\textbf{Tasks}} \\
    \cline{3-4}
    \multirow{-2}{*}{\textbf{Methods}} & \multirow{-2}{*}{\textbf{Complexity}} & 
    \textbf{Pair Node Cls} & 
    \textbf{Link Pred} \\ 
    \midrule  
    Degree & \color{black}{${O\big( |\mathcal{V}| \big)}$} & $0.930 \pm 0.012$ & $0.830 \pm 0.023$ \\ 
    Betweenness & $O\big( |\mathcal{V}|^3 \big)$ & $0.944 \pm 0.007$ & $0.756 \pm 0.031$ \\ 
    Harmonic & $O\big( |\mathcal{V}|^3 \big)$ & $0.958 \pm 0.018$ & $0.733 \pm 0.032$ \\
    Closeness & $O\big( |\mathcal{V}|^3 \big)$ & $0.978 \pm 0.007$ & $0.765 \pm 0.012$ \\
    Load & $O\big( |\mathcal{V}|^3 \big)$ & $0.974 \pm 0.012$ & $0.778 \pm 0.031$ \\ 
    \midrule 
    P-GNN~\cite{you2019position} & \color{black}{${O\big( |\mathcal{E}| + |\mathcal{V}| \log |\mathcal{V}| \big)}$} & $0.892 \pm 0.093$ & $0.565 \pm 0.040$ \\
    PSGNN & \color{black}{${O\big( |\mathcal{V}| \big)}$$^{\dagger}$} & \color{red}{$\mathbf{0.997 \pm 0.007}$} & \color{red}{$\mathbf{0.963 \pm 0.007}$}\\
    \bottomrule 
    \end{tabular}
    }
    \label{table:heuristic}
\end{table}

\section{Experiment}
\subsection{Datasets and Tasks}
We conduct experiments on both synthetic and real-world graph datasets. We employ the following three datasets for the link prediction task.

\noindent
\textbf{Communities}. A synthetic connected caveman graph with 20 communities, each community consisting of 20 nodes. Every pair of communities are mutually isomorphic. 

\noindent
\textbf{Grid}. A synthetic 2D grid graph of size $20 \times 20$ (400 nodes) with all the nodes having the same feature. 

\noindent
\textbf{Bio-Modules}~\cite{espinosa2010specialization}. An artificial gene regulatory network produced by simulated evolutionary processes. A link reflects two genes have mutual effects. Each node contains a two-dimensional one-hot vector, indicating the corresponding gene is either in activation or repression.

\noindent
\textbf{Cora}, \textbf{CiteSeer} and \textbf{PubMed}~\cite{velivckovic2017graph}. Three real-world citation networks consisting of 2708, 3327, and 19717 nodes, respectively, with each node representing an artificial intelligence article. Every dimension of the node's feature vector is a binary value, indicating the existence of a word. 

We also perform the pairwise node classification task on the following three datasets. In pairwise node classification, every two nodes are associated with a label. We are predicting the labels of a group of unseen node pairs. 
\begin{itemize}
    \item \textbf{Communities}. The structure is the same as described above. Each node's label corresponds to the community it is belonging.

    \item \textbf{Emails}. 
Seven real-world email communication networks from SNAP~\cite{leskovec2007graph}. Each graph comprises six communities. Every node has a label indicating the community to which it belongs. 
Each node represents a person. A link between person $u$ and person $v$ indicates that person $u$ has sent at least one email to person $v$. 

    \item \textbf{Protein}. 1113 real-world protein graphs from~\cite{borgwardt2005shortest}. Each node contains a 29-dimensional feature vector and is labeled with a functional role of the protein. 
\end{itemize}

\subsection{Experimental Setups}
\noindent 
\textbf{Networks}: We employ the Generalized graph convolution network~\cite{li2020deepergcn} as the network of the anchor selection component. 
The embedding size of both an anchor as well as a vertex is set to be the same. In our implementation, this number is 32. This setting is consistent throughout all the evaluated models. 

\noindent
\textbf{Train/Valid/Test Split}. 
We follow the setting from~\cite{you2019position} and use 80\% existing links and the same number of non-existing ones for training, as well as the remaining two sets of 10\% links for validating and testing, respectively. We report the performance on the test set using the model which achieves the highest accuracy on the validation set. For a task on a specific dataset, we independently train ten models with different random seeds and train/validation splits, and report the average performance and the standard deviation. 
As for the choice of hyperparameters, we follow the same setting as~\cite{you2019position}. We have also evaluated the models with various combinations of hyperparameters. The performance did not show significant difference with the models using the hyperparameters of~\cite{you2019position}. 

\noindent
\textbf{Loss Functions}. Binary cross-entropy is employed as the loss function to train the PSGNN.

\noindent
\textbf{Baselines}.
We compare the PSGNN against existing popular GNN models. To ensure the reliability and reproducibility of the experimental results, we carefully selected and adjusted the hyperparameters for each model.

\noindent
\textbf{Hyperparameters}.

For each model, we conducted a grid search on the validation set to tailor the optimal hyperparameters specifically for the pairwise node classification on the communities dataset and the link prediction on the communities dataset. We present these hyperparameters in \autoref{tab:hyperparameter}. For other datasets in the pairwise node classification, we applied the hyperparameters optimized for the communities dataset. Similarly, for other datasets in the link prediction, we utilized the hyperparameters fine-tuned for the communities dataset. Next, we will provide detailed explanations of some hyperparameters, based on experiments conducted on the pairwise node classification and link prediction tasks using the communities dataset.

We uniformly set the number of training epochs to 2000 for all models. Through experimentation, we have determined that 2000 epochs are sufficient for all the models under consideration to reach convergence. To avoid overfitting, we adopt an early stopping strategy and select the model with the best performance on the validation set.

We utilize the Adam~\cite{kingma2014adam} optimizer for training all the models. Adam is well-suited for this task as it leverages both the moving averages of the gradients and the element-wise square of the gradients to dynamically adjust the learning rates during the training process. This adaptive learning rate mechanism can compensate for suboptimal choices of the initial learning rate. Through our experiments, we have established that, between the Adam and SGD~\cite{robbins1951stochastic}, Adam is the better choice.

Next, I will show the search ranges for different hyperparameters used in the grid search.

\begin{itemize}
    \item{Learning rate:} We searched starting from $0.0005$ and doubling it each time (i.e., $0.0005$, $0.001$, $0.002$, etc.), continuing this process up to $0.064$. Because we found that the AUC score did not improve with the increase in learning rate, we stopped when we searched for $0.064$.
    \item{Dropout rate:} We searched starting from $0.0$ and incrementing by $0.1$ each time (i.e., $0.0$, $0.1$, $0.2$, etc.), continuing this up to $0.9$.
    \item{The number of heads for GAT:} We searched starting from $1$ and incrementing by $1$ each time (i.e., $1$, $2$, $3$, etc.), continuing this up to $5$. Because we found that the AUC score did not improve with the increase in the number of heads for GAT, we stopped when we searched for $5$.
    \item{The number of layers:} We searched starting from $2$ and incrementing by $1$ each time (i.e., $2$, $3$, $4$, etc.), continuing this up to $5$. Layers includes input layer, output layer, and hidden layers. Because we found that the AUC score did not improve with the increase in the number of layers, we stopped when we searched for $5$.
    \item{The dimension of node embedding:} We searched starting from $1$ and doubling it each time (i.e., $1$, $2$, $4$, etc.), continuing this process up to $512$. Because we found that the AUC score did not improve with the increase in the dimension of node embedding, we stopped when we searched for $512$.
    \item{Batch normalization:} We searched use both using and not using it.
\end{itemize}

Furthermore, We set the number of anchors to be $\log N$, where $N$ is the number of nodes within a graph. These settings follow~\cite{you2019position}. Drawing from the Bourgain theorem presented in~\cite{you2019position}, we understand that choosing $\log^2 N$ anchors can guarantee low distortion embedding. However, our experiments have revealed that opting for $\log N$ anchors also yields similar performance while reducing computational cost. To show it, we conducted a series of experiments, the results of which are showcased in \autoref{table:logn}.

$\alpha$ in \autoref{eq:o_epsilon} is set to be $0.5$. 
All the GNN models contain three layers. The implementation is based on the PyTorch framework trained with two NVIDIA 2080-Ti GPUs.

\begin{table*}[tb]
\caption{Optimal hyperparameter identified through hyperparameter tuning.}
\centering
\begin{subtable}[t]{0.8\textwidth}
\caption{Optimal hyperparameter on pairwise node classification.}
\resizebox{1.0\textwidth}{!}{
\begin{tabular}{l c c c c c c c c}
\toprule
Methods & Number of & Learning  & Dropout & Number of & Dimension of & The Number & Number of & Batch  
\\
& Training Epochs & Rate & Rate & Anchors & Node Embedding & of Heads & Layers & Normalization
\\ 
\midrule
GCN~\cite{kipf2016semi} & 2000 & 0.008 & 0.5 & NA & 64 & NA & 3 & Used\\
GraphSAGE~\cite{hamilton2017inductive} & 2000 & 0.002 & 0.5 & NA & 64 & NA & 3 & Used \\
GAT~\cite{velivckovic2017graph} & 2000 & 0.004 & 0.5 & NA & 64 & 2 & 4 & Used \\
GIN~\cite{xu2018powerful} & 2000 & 0.004 & 0.6 & NA & 32 & NA & 3 & Used \\

\midrule 
\multicolumn{9}{c}{Position-Aware Methods} \\
\midrule 
P-GNN~\cite{you2019position} & 2000 & 0.001 & 0.4 & $\log N$ & 128 & NA & 4 & Unused \\
A-GNN~\cite{liu2020graph} & 2000 & 0.001 & 0.2 & $\log N$ & 64 & NA & 3 & Used \\
PSGNN & 2000 & 0.001 & 0.3 & $\log N$ & 128 & NA & 4 & Unused \\
\bottomrule
\end{tabular}
}
\end{subtable}
\label{tab:asd1}

\begin{subtable}[t]{0.8\textwidth}
\caption{Optimal hyperparameter on link prediction.}
\resizebox{1.0\textwidth}{!}{
\begin{tabular}{l c c c c c c c c}
\toprule
Methods & Number of & Learning  & Dropout & Number of & Dimension of & The Number & Number of & Batch
\\
& Training Epochs & Rate & Rate & Anchors & Node Embedding & of Heads & Layers & Normalization
\\
\midrule
GCN~\cite{kipf2016semi} & 2000 & 0.008 & 0.4 & NA & 64 & NA & 3 & Used \\
GraphSAGE~\cite{hamilton2017inductive} & 2000 & 0.002 & 0.6 & NA & 64 & NA & 3 & Used \\
GAT~\cite{velivckovic2017graph} & 2000 & 0.004 & 0.5 & NA & 64 & 2  & 4 & Used \\
GIN~\cite{xu2018powerful} & 2000 & 0.004 & 0.6 & NA & 64 & NA & 3 & Used \\

\midrule 
\multicolumn{9}{c}{Position-Aware Methods} \\
\midrule 
P-GNN~\cite{you2019position} & 2000 & 0.001 & 0.3 & $\log N$ & 128 & NA & 4 & Unused \\
A-GNN~\cite{liu2020graph} & 2000 & 0.001 & 0.4 & $\log N$ & 128 & NA & 3 & Used \\
PSGNN & 2000 & 0.001 & 0.3 & $\log N$ & 128 & NA & 4 & Unused \\
\bottomrule
\end{tabular}
}
\end{subtable}
\label{tab:hyperparameter}
\end{table*}

\begin{table}[t]
    \caption{AUC scores and training time of PSGNN model across task pairwise node classification and link prediction on various databases: impact of number of anchors on AUC and training time (seconds), using two NVIDIA 2080-Ti GPUs.}
    \centering
    \resizebox{0.45\textwidth}{!}{
    \begin{tabular}{c c c c}
    \midrule 
    \textbf{Databaset} & \textbf{Number of Anchors} & \textbf{AUC scores} & \textbf{Time} \\
    \midrule 
    \multicolumn{4}{c}{Task: pairwise node classification} \\
    \midrule 
    Communities & $\log^2 N$ & $0.979$ & $12.29$ \\
    Communities & $\log N$ & $0.978$ & $8.92$ \\
    \midrule
    Email & $\log^2 N$ & $0.739$ & $84.25$ \\
    Email & $\log N$ & $0.736$ & $60.32$ \\
    \midrule
    Protein & $\log^2 N$ & $0.541$ & $4014.71$ \\
    Protein & $\log N$ & $0.533$ & $2874.59$ \\
    \midrule 
    \multicolumn{4}{c}{Task: link prediction} \\
    \midrule 
    Communities & $\log^2 N$ & $0.955$ & $14.31$ \\
    Communities & $\log N$ & $0.954$ & $10.07$ \\
    \midrule
    Grids & $\log^2 N$ & $0.947$ & $12.98$ \\
    Grids & $\log N$ & $0.942$ & $10.18$ \\
    \midrule
    Bio-Modules & $\log^2 N$ & $0.990$ & $659.48$ \\
    Bio-Modules & $\log N$ & $0.989$ & $430.35$ \\

    \bottomrule 
    \end{tabular}
    }
    \label{table:logn}
\end{table}

\subsection{Comparison Against State-of-the-Art}
\autoref{table:compare_with_sota} demonstrate comparing the performance of PSGNNs with existing competitive GNN models on the above datasets for pairwise node classification and link prediction. We discover that since the GCN~\cite{kipf2016semi} and its variants cannot be aware of node positions, they reveal nearly random performance on both tasks' datasets. Other position-unaware models show negligible improvements over GNNs. In contrast, P-GNNs exhibits breakthrough improvements on some datasets. Moreover, our PSGNNs further substantially enhance the performance over P-GNNs, achieving the best AUC scores on all the datasets of both tasks. In particular, the average increase on the email and grid datasets reaches more than 27\%, and the AUC scores for all the synthetic datasets surpass 0.9. These performances indicate the effectiveness of our PSGNNs. 

On the other hand, PGNNs first select different anchor sets, then within each anchor set, a node chooses the anchor as the one having the minimum distance to the node within the set. In contrast, instead of selecting anchors from the pre-selected anchor sets as P-GNNs, PSGNNs directly select anchors, leading to a lower computational complexity than that of P-GNNs.

\subsection{Analysis on Scalability}
We also study the scalability of PSGNNs and other GNNs. That is, how the performance changes as the graph size increases. The results are reported in \autoref{tab:size_analysis}. We discover that for the GNNs to be position-unaware, their performances are similar to randomness, regardless of the graph size.  

\blfootnote{$^{\dagger}$Complexity of PSGNN does not consider calculating pairwise node distances happening in the preprocessing stage. }
In contrast, for pairwise node classification on communities, both GNNs can learn the relative positions among nodes, \ie PGNNs and PSGNNs, show increasing AUC scores as the graph size increases. Besides, PSGNNs demonstrate better performance for all the graph sizes over PGNNs and better stability. Moreover, the PSGNNs performance steadily grows, whereas the trend of PGNNs depicts fluctuations. On the other hand, for link prediction on grids, the PGNNs scalability is limited, revealing a decreasing tendency with the increasing graph size. On the contrary, PSGNNs still maintain a positive correlation between AUC and the graph sizes, indicating their general flexibility on different datasets and tasks.

\subsection{Necessity of Learning Anchors}

We have described before that selecting proper anchors is of NP-completeness. Although it is impractical to discover an optimal solution toward an NP-complete problem, rule-based heuristic algorithms can lead to sub-optimal approaches. For example, \cite{liu2020graph} proposes to select the nodes with the most significant degrees as anchors. Nonetheless, although these rule-based heuristic methods have lower computational complexity than the exponential level, the run-time is still higher than the linear cost of our anchor-learning component (see \autoref{table:heuristic} for the comparison in detail). 
Specifically, 
PSGNN uses a standard GNN to assign the likelihoods of being selected as anchors, 
and selecting one of the top-$k$ anchors can be done in linear time. That is, PSGNN takes $O(|\mathcal{V}|)$ to select anchors. 
Thus, the rule-based approaches have more insufficient scalability than PSGNNs. Furthermore, we also compare the performance of learning anchors against rule-based selections on both pairwise node classification and link prediction tasks. 
The datasets for the two tasks are a connected Caveman graph with 64 communities, each having 64 nodes, and a $48 \times 48$ grid network, respectively. 
We observe that the PSGNN consistently outperforms the rule-based heuristic methods while it enjoys a lower computational complexity.

\section{Limitations}
We describe limitations below. 

\textbf{Complexity}: Although happening in the preprocessing stage, it still requires $O(|\mathcal{V}|^3)$ ($|\mathcal{V}|$ is the number of nodes) time complexity to calculate the pairwise node distances. Hence, it requires further investigation on the scalability of the proposed approach over very giant graphs. 

\section{Conclusion}
Despite increasing attention in GNNs, relatively little work focuses on predicting the pairwise relationships among graph nodes. Our proposed position-sensing graph neural networks (PSGNNs) can extract relative positional information among graph nodes, substantially improving pairwise node classification and link prediction over a variety of baseline GNNs. Our PSGNN also exhibit great scalability. 
Apart from practical contributions, theoretically, this paper brings NP-completeness theory in analyzing the complexity of GNN algorithms.

\vspace{.5em}
\small{
\noindent \textbf{Acknowledgments}: 
This work was partly supported by the National Research Foundation of Korea(NRF) grant funded by the Korea government(MSIT) (No. 2020R1F1A1061667).
}

\ifCLASSOPTIONcaptionsoff
  \newpage
\fi

\bibliographystyle{IEEEbib}
\bibliography{refs}

\end{document}